\documentclass[letterpaper, 10 pt, conference]{ieeeconf}
\IEEEoverridecommandlockouts                              

\usepackage{graphics} 
\usepackage{amssymb}  

\usepackage{cite}
\usepackage{amsmath,amssymb,amsfonts}
\usepackage{graphicx}
\usepackage{textcomp}
\usepackage{xcolor}
\def\BibTeX{{\rm B\kern-.05em{\sc i\kern-.025em b}\kern-.08em
    T\kern-.1667em\lower.7ex\hbox{E}\kern-.125emX}}

\usepackage[hyphens]{url}
\usepackage[hidelinks]{hyperref}
\hypersetup{breaklinks=true}

\usepackage{xspace}
\usepackage{multirow}
\usepackage{booktabs} 
\usepackage{amsfonts}
\usepackage{amsmath}

\usepackage{algorithm}
\usepackage{mathtools}
\usepackage[noend]{algpseudocode}
\usepackage{color}
\usepackage{array}
\usepackage{pgfplots}

\usepackage{color,soul}

\usepackage[export]{adjustbox}
\usepackage{multirow}

\usepackage[caption=false]{subfig}

\usepackage{lipsum} 

\usepackage{setspace}

\usepackage{wrapfig}

\algdef{SE}[DOWHILE]{Do}{doWhile}{\algorithmicdo}[1]{\algorithmicwhile\ #1}%

\title{\LARGE \bf
Rethinking Energy Management for Autonomous Ground Robots on a Budget}

\author{Akshar Chavan$^{1}$, Rudra Joshi$^{1}$ and Marco Brocanelli$^{1}$ 
\thanks{*This work was supported by NSF award CNS-1948365}
\thanks{$^{1}$Akshar Chavan, Rudra Joshi and Marco Brocanelli  are with the Energy Aware Systems Laboratory (EAS), at The Ohio State University, OH 43210, USA
        {\tt\small \{chavan.43, joshi.485, brocanelli.1\}@osu.edu}}%
}

\usepackage{tcolorbox}

\algnewcommand\algorithmicinput{\textbf{Input:}}
\algnewcommand\INPUT{\item[\algorithmicinput]}
\algnewcommand\algorithmicoutput{\textbf{Output:}}
\algnewcommand\OUTPUT{\item[\algorithmicoutput]}

\pgfplotsset{compat=1.18}
\begin{document}

\maketitle
\thispagestyle{plain}
\pagestyle{plain}

\begin{abstract}

Autonomous Ground Robots (AGRs) face significant challenges due to limited energy reserve, which restricts their overall performance and availability. Prior research has focused separately on energy-efficient approaches and fleet management strategies for task allocation to extend operational time.
A fleet-level scheduler, however, assumes a specific energy consumption during task allocation, requiring the AGR to fully utilize the energy for maximum performance, which contrasts with energy-efficient practices.
This paper addresses this gap by investigating the combined impact of  computing frequency and locomotion speed on energy consumption and performance. We analyze these variables through experiments on our prototype AGR, laying the foundation for an integrated approach that optimizes cyber-physical resources within the constraints of a specified energy budget. To tackle this challenge, we introduce PECC (Predictable Energy Consumption Controller), a framework designed to optimize computing frequency and locomotion speed to maximize performance while ensuring the system operates within the specified energy budget.
We conducted extensive experiments with PECC using a real AGR and in simulations, comparing it to an energy-efficient baseline. 
Our results show that the AGR travels up to 17\% faster than the baseline in real-world tests and up to 31\% faster in simulations, while consuming 95\% and 91\% of the given energy budget, respectively. 
These results prove that PECC can effectively enhance AGR performance in scenarios where prioritizing the energy budget outweighs the need for energy efficiency.
\looseness -1


\end{abstract}

\vspace{-1mm}
\section{INTRODUCTION}
\label{sec:intro}
\vspace{-1mm}
In recent years, extensive research has been conducted on Autonomous Ground Robots (AGRs) due to their ability to autonomously carry out a diverse range of tasks, including face recognition~\cite{YU2021107128}, security breach detection~\cite{AMR_patrol3}, and delivery services~\cite{Srinivas_Ramachandiran_Rajendran_2022, KAMRa_Ayanian_2015, Heimfarth_Ostermeier_Hübner_2022}.

To enable these complex functions, AGRs are equipped with advanced cyber-physical components. 
However, these components heavily rely on limited-capacity rechargeable batteries as their primary energy source. 
Despite technological advancements, the energy density of these batteries has not kept pace with the increasing demands of mobile systems, posing challenges in maximizing AGRs' performance.
To address this problem, research efforts are directed into two main directions. Firstly, providing AGR fleet task and charging allocation algorithms that maximize utilization of the AGRs within the available onboard energy \cite{ tcm, 9981285, 9636815, Boccia_Masone_Sterle_Murino_2023, Chen_Xie_2022, 7294146, McGowen_Dagli_Dantam_Belviranli_2024, tcm_m, Atik_Chavan_Grosu_Brocanelli_2023}. 

These algorithms account for energy availability, tasks to perform, and the energy required for execution. 
They use energy models to predict the energy required for tasks over a specified period. Consequently, they allocate tasks to AGR for a certain period of time, indirectly defining a task energy budget.  \looseness -1
 
Secondly, minimizing the energy consumption of AGR cyber-physical components, with particular emphasis on computing resources (cyber) and locomotion (physical), which account for a significant portion of the overall energy consumption~\cite{e2m}. These approaches use techniques like Dynamic Voltage and Frequency Scaling (DVFS) and other similar methods to minimize computing energy~\cite{e2m, Rossi_Vaquero_Sanchez_Net_da_Silva_Vander_Hook_2020, Tang_Shah_Michmizos_2019}. For locomotion, they employ strategies such as energy-efficient path planning~\cite{Lamini_Benhlima_Elbekri_2018, Kim_Kim_2014, Li_Wang_Chen_Kan_Yu_2023, Di_Franco_Buttazzo_2015, 6385568} and develop hardware components with lower energy demands~\cite{Kim_Shin_Lee_Lee_Yoo_2018, Frank_Yang_Liu_2021}.
Primarily, these studies focus independently on minimizing either computational or locomotion energy consumption. However, the total energy consumption and performance of the AGR depends on the combined utilization of cyber-physical resources~\cite{eemrc}. 
Considering all the above research, it is worth noting that defining task-specific or time-based energy budgets and minimizing energy consumption do not always align seamlessly. \looseness -1

While performing tasks, AGRs navigate through obstacles, which can result in high variability in energy consumption than initially estimated by the task allocation algorithm, even if they are energy efficient. Additionally, when an AGR is assigned a specific energy budget by a fleet manager, prioritizing energy efficiency can sometimes unintentionally limit performance. This is because energy-efficient solutions may require lowering  computing frequency and locomotion speed, as detailed in Section~\ref{sec:exp_results}. Moreover, the operational environment of AGRs can create significant discrepancies between actual energy consumption and the estimated budget. These discrepancies pose critical challenges for accurate task allocation and resource management, necessitating adjustments to schedules based on updated energy availability for all AGRs. Such adjustments can potentially impact the overall fleet performance~\cite{tcm, tcm_m}. 
\looseness -1

Given that allocation algorithms assign an energy budget for the scheduled tasks to maximize the performance and utilization of an AGR fleet, \textit{we argue that it is logical to focus on maximizing the consumption of this allocated budget for the specified tasks rather than prioritizing energy efficiency at the expense of performance,  such as travel time.} For example,
in a food delivery task, maximizing speed within a specified energy budget can enhance customer satisfaction by reducing delivery time. To make the most of the allocated energy budget, it is thus essential to implement an energy management system on each AGR that ensures \emph{a predictable use of energy} for maximized performance within the allocated energy budget. \looseness -1

To overcome the above challenges, this paper proposes the Predictable Energy Consumption Controller (\textsf{PECC}), which adjusts computing frequency and locomotion speed to minimize the AGR travel time during task execution while \textit{predictably} consuming a given energy budget for the task.  This paper's specific contributions are: \looseness -1

\begin{itemize}
    \item We conduct extensive experiments with our prototype AGR to study its performance across various CPU frequencies and maximum speed.
    \item We formulate the PECC optimization problem to maximize the energy  budget utilization and task performance by controlling frequency and speed, and test it on our prototype AGR as well as through simulation results against an energy-efficient baseline.
    \item In experimental results, the AGR travels up to 17\% faster than the baseline in real-world tests and up to 31\% faster in simulations, while consuming 95\% and 91\% of the given energy budget, respectively. \looseness -1
       
\end{itemize}

The rest of the paper is organized as follows. Section~\ref{sec:backgraound_and_motivation} describes the motivation. Section~\ref{sec:design} details the design  of \textsf{PECC}. Section~\ref{sec:exp_results} presents the experimental results. Section~\ref{sec:related} studies the related work. Finally,  Section~\ref{sec:conclusion} concludes the paper. \looseness -1

\vspace{-3mm}
\section{Motivation}
\label{sec:backgraound_and_motivation}
\begin{figure}[!t]
  \centering
      \subfloat[]{
  
  \includegraphics[width=0.21\textwidth]{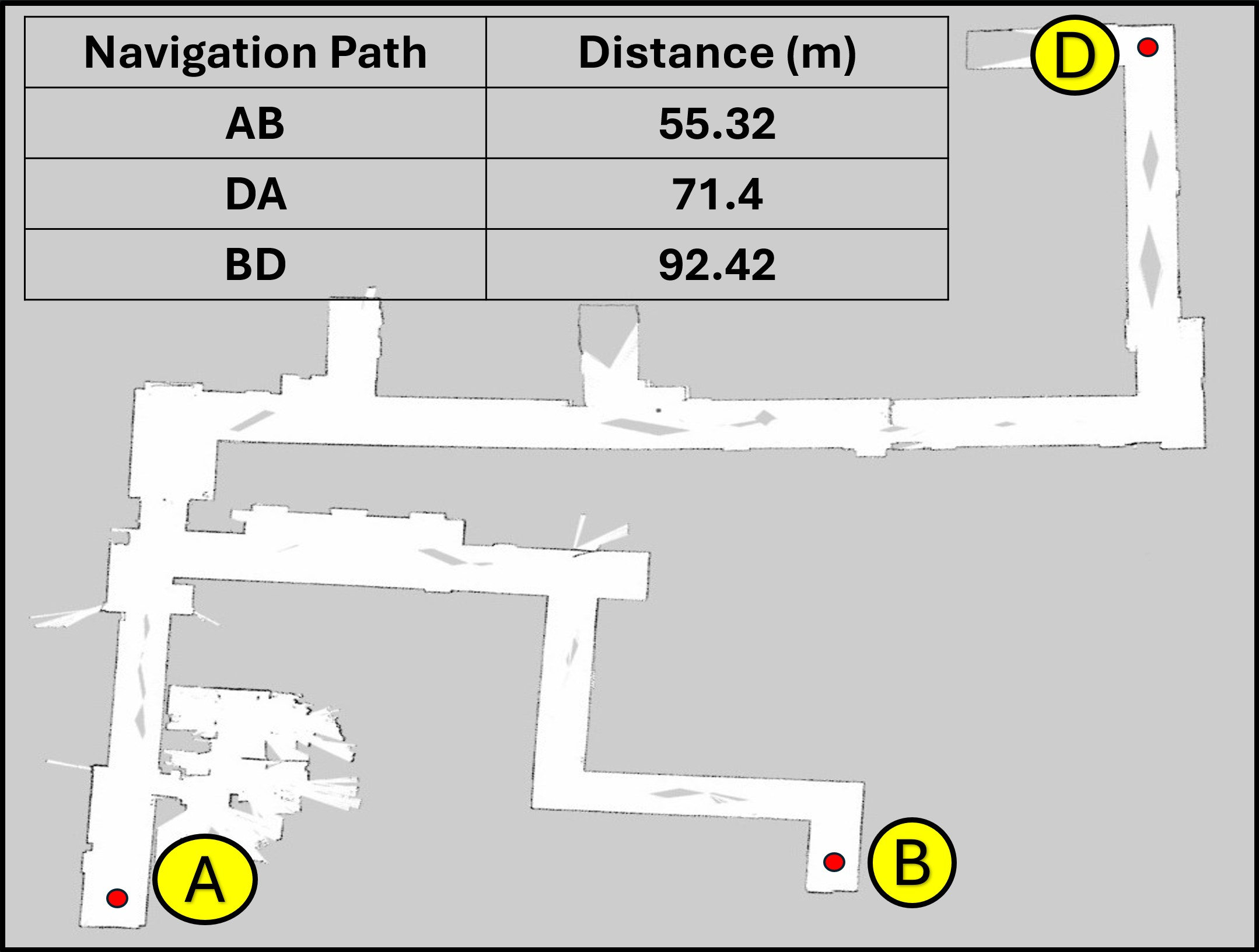}
  \label{fig:real_map}
  }
    \subfloat[]{
  \includegraphics[width=0.21\textwidth]{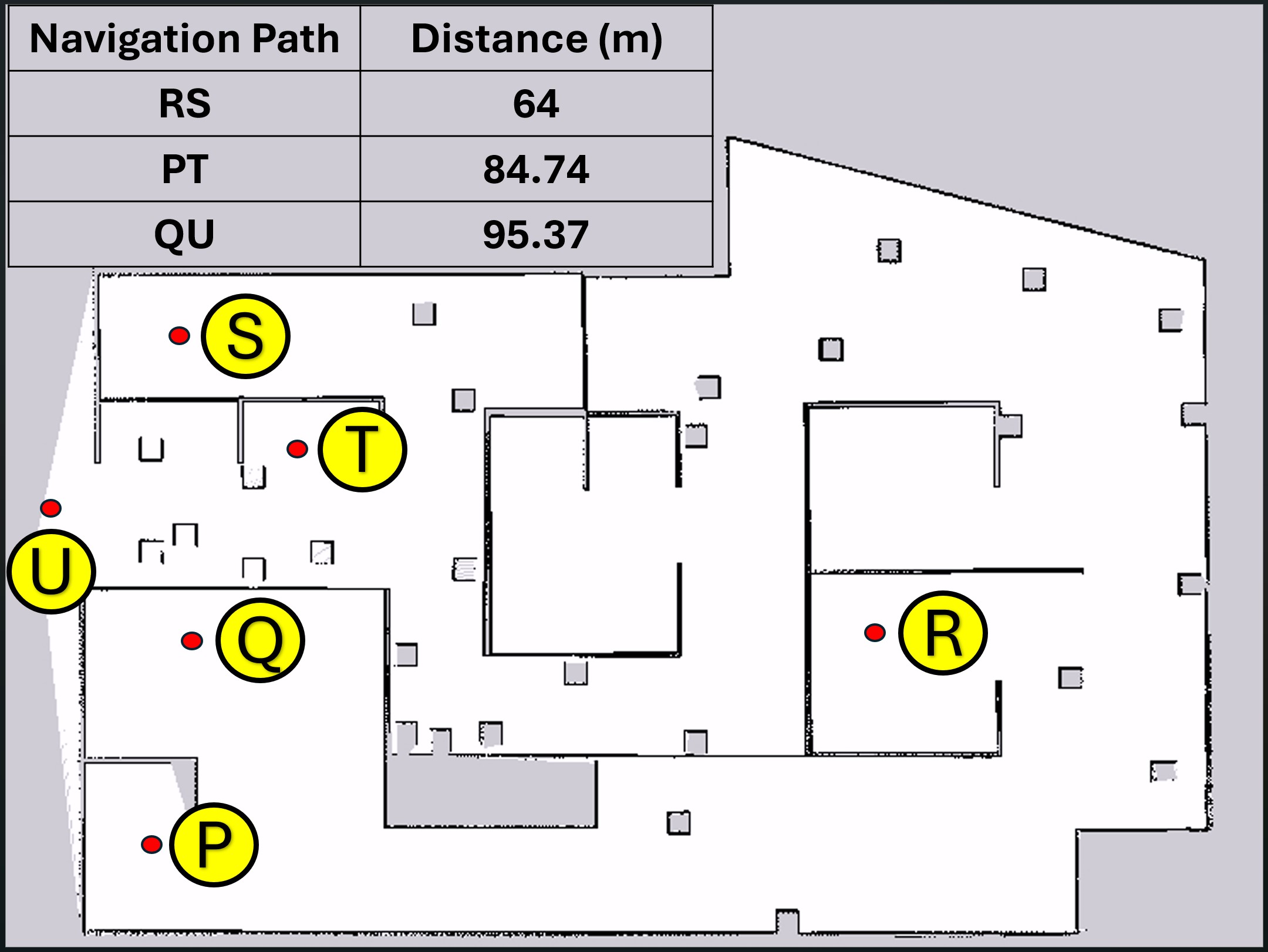} 
  
  \label{fig:sim_map}
  }
 \\
 \vspace{-0.3cm}
\subfloat[]{
  \includegraphics[width=0.21\textwidth, trim=0 0 0 0, clip]{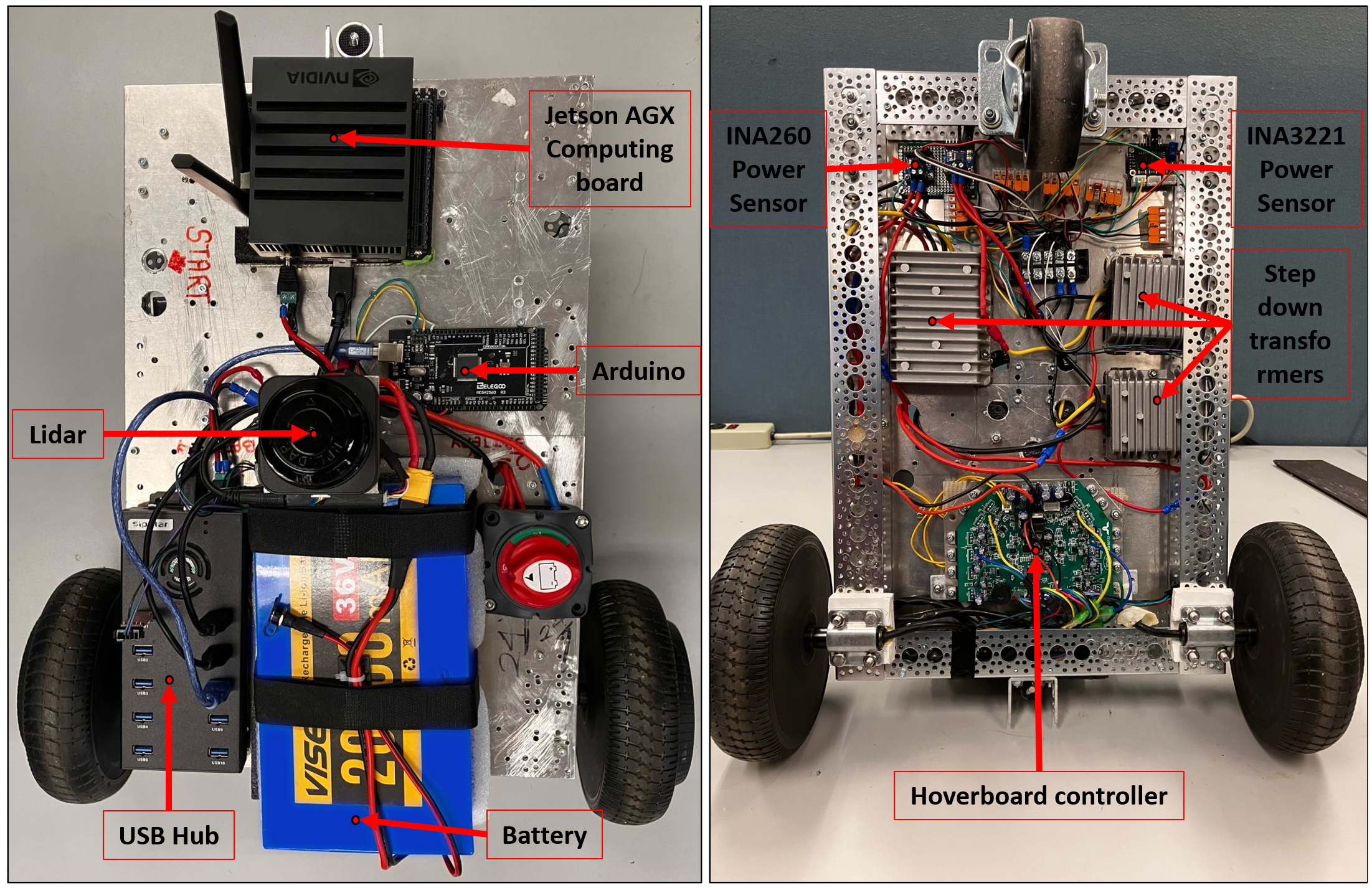}
  \label{fig:robot}
  }
          \subfloat[]{  
  \includegraphics[width=0.21\textwidth, trim=0 40 0 60, clip]{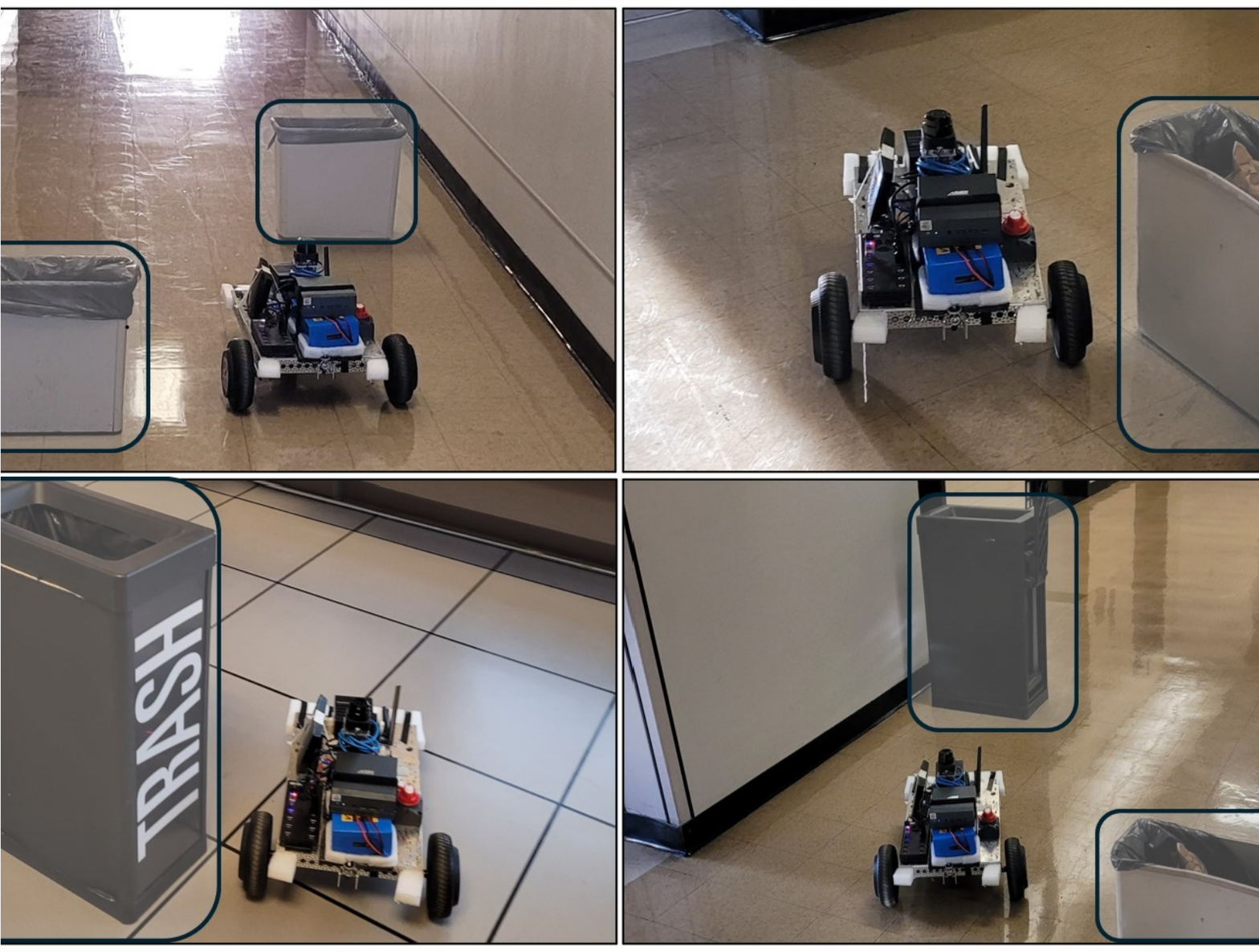}
  \label{fig:agr_on_path}

  }
  \vspace{-1mm}
\caption{Experimental setup used in our experiments: (a) real-world map, (b) map for simulations, (c) our prototype AGR, and (d) example obstacles in the real map.}
  \label{fig:map_and_amr}
  \vspace{-0.7cm}
\end{figure}

\vspace{-1.3mm}
In this section we first explain the hardware and software components of the AGR shown in Figure~\ref{fig:robot}, which is used for conducting experiments, followed by a description of the experiment setup. Then, we present our observations from the experimental data to motivate our research goal.


\noindent{\textbf{Hardware, Software, and Navigation Stack.}} The platform used in our experiments comprises typical components of modern AGRs, including an NVIDIA Jetson AGX Xavier for computational tasks, INA3221 and INA260 Power Sensors for measuring power consumption of sensors, computing, and locomotion components, a Slamtec RPLidar S1 LiDAR, a hoverboard motor controller with two 250W hoverboard motors for locomotion, and a 36V 20Ah Li-ion battery to power the system. 
To meet the maximum current requirement of the components, we used three step-down transformers. The AGR runs ROS Melodic on Ubuntu 18.04. \looseness -1

We generated a map of our laboratory environment using the ROS GMapping package and defined a set of predetermined routes with specified starting and ending points, shown in Figure~\ref{fig:real_map}. For navigation, we employed the ROS-based stack, specifically incorporating the MoveBase package and the Dynamic Window Approach (DWA) planner. We selected DWA due to its widespread use in robotics research~\cite{dwa1, dwa2, dwa3, dwa4, dwa5, Liu_Zhong_Willcock_Fisher_Shi_2023}, as well as the ongoing efforts to enhance its application in ground robots. 
The MoveBase package integrated global and local path planning for efficient navigation. The global planner, Navfn, uses Dijkstra's algorithm for optimal paths, while the DWA planner handles real-time obstacle avoidance and trajectory optimization based on the robot's constraints.
\looseness -1

\noindent{\textbf{Profiling Energy Consumption and Performance.}} 
We used our AGR to collect data on various parameters, including speed, power consumption, and end-to-end latency (which encompasses the time to acquire sensor data, analyze it, and react to the detected obstacles by updating the speed and direction of the AGR) for different CPU frequency~$f$ and \textit{maximum achievable speed}~$s_{max}$ pairs. The term $s_{max}$ represents the configurable speed limit of the AGR, while the actual speed~$s \leq s_{max}$ is determined by the DWA planner. The frequency~$f$ varied from 1.11GHz to 2.26GHz, and the maximum achievable speed~$s_{max}$ ranged from 0.4m/s to 2.4m/s with increments of 0.4m/s. We limited the minimum CPU frequency to 1.11GHz because for lower frequencies the AGR is unable to move correctly, specially at high~$s_{max}$ due to the excessively high end-to-end latency. For each~$(f, s_{max})$ pair, we conducted five runs and averaged the results to profile the power consumption and end-to-end latency of the AGR.  \looseness -1


\vspace{-2mm}
\subsection{Motivation}

\begin{figure*}[!t]
\vspace{-2mm}
  \centering
      \subfloat[]{
  
  \includegraphics[width=0.205\textwidth]{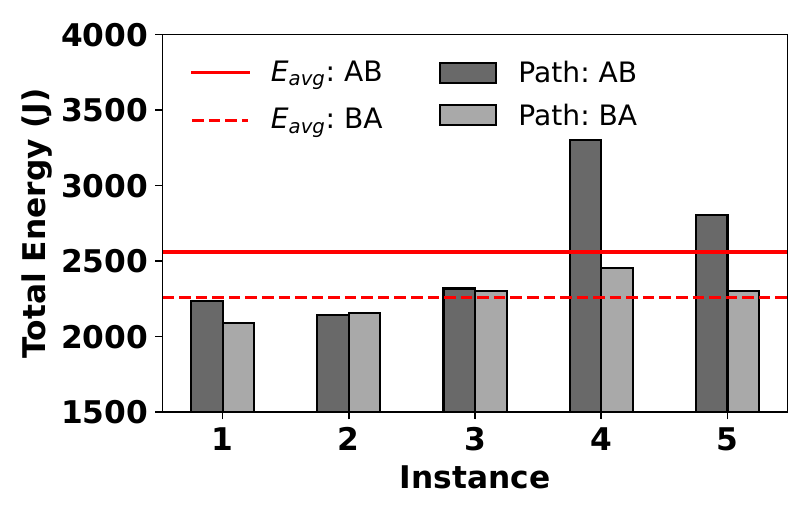}
  \label{fig:total_en}
  }
      \subfloat[]{
  \includegraphics[width=0.205\textwidth]{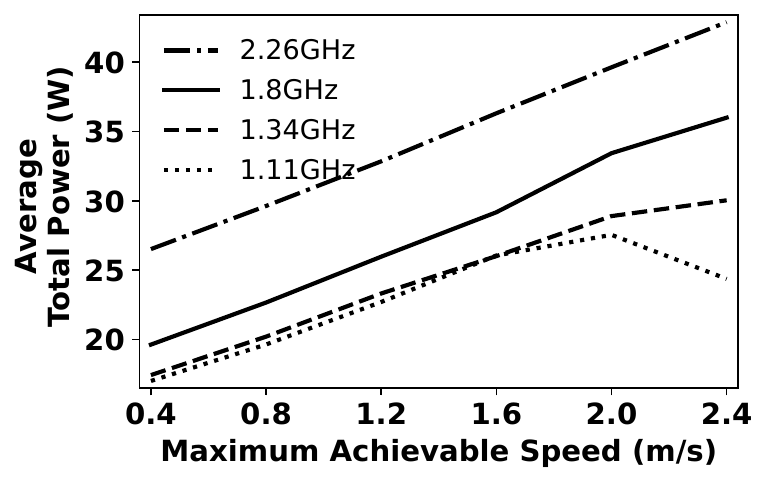}
  
  \label{fig:total_power}
  }
  \subfloat[]{
  \includegraphics[width=0.205\textwidth]{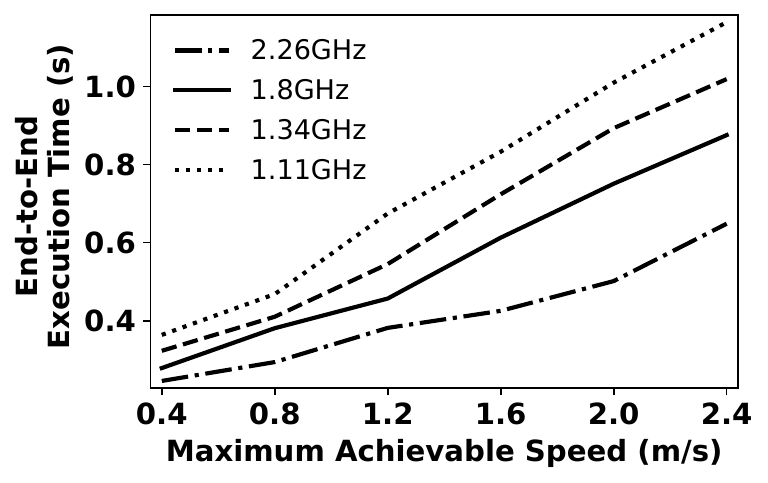}
  \label{fig:e2e}
  }
      \subfloat[]{
  \includegraphics[width=0.205\textwidth]{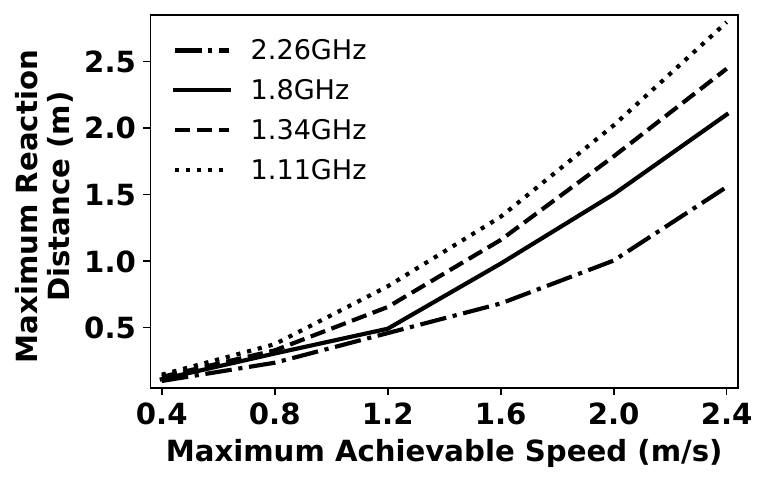}
  \label{fig:rd}
  }
\vspace{-2mm}
  \caption{Motivation results: (a) Total energy consumed by the AGR while navigating the paths AB and BA on the map shown in Figure~\ref{fig:real_map}, (b) average power consumption, (c) end-to-end execution time, and (d) maximum reaction distance at different computing frequencies and maximum achievable speed. \looseness -1 }
  \label{fig:f_v_pair}
  \vspace{-0.7cm}
\end{figure*}

As discussed in Section \ref{sec:intro}, obstacles encountered during navigation influence the AGR's energy consumption by requiring speed adjustments.
Consequently, the total energy consumed by an AGR can differ substantially depending on the complexity of the environment. Figure~\ref{fig:total_en} highlights this variation in the total energy consumed by the AGR across five experimental runs while traversing the navigation path from position A to position B (path AB) and from position B to position A (path BA) on the map shown in Figure~\ref{fig:real_map}~(Figure ~\ref{fig:sim_map} shows the map used for simulations in Sections~\ref{sec:exp_results}). The experiment was conducted using the energy-efficient approach, which aims to minimize the energy per meter~(J/m) by selecting an energy-efficient  ($f,s_{max}$) pair for the AGR. We describe this approach in detail in Section \ref{sec:exp_results}.  \looseness -1

Figure~\ref{fig:total_en} shows that the average energy consumption for path AB is 2560J with a standard deviation of 434J, while for path BA, it is 2261J with a standard deviation of 127J. 
This variation  across runs is due to new obstacles encountered in instances 4~and~5 (see example in Figure~\ref{fig:agr_on_path}), which were not present in the map used for instances 1~to~3. In instances 1~to~3, the AGR traversed paths AB and BA with an average energy consumption of 2232J and 2183J, respectively. In instances 4~and~5, the introduction of seven additional obstacles  necessitated path  replanning, leading to increased energy consumption of 3051J for path AB and 2378J for path BA. 
This experiment shows that energy consumption varies with environmental complexity. Such variability is difficult to anticipate at task allocation time. Therefore, the AGR should manage the energy budget usage during task execution.
\looseness -1 

\noindent{\textbf{Cyber-Physical Energy Management.}}
We treat the AGR as a cyber-physical system, recognizing that interactions between locomotion and computing impact performance and energy consumption~\cite{eemrc}. 
For instance, lowering the CPU frequency to conserve computing energy increases the latency in updating the speed commands (\textit{cmd\_vel}) for a perceived environment. Delays in \textit{cmd\_vel} causes the AGR's locomotion controller to use a  defaults speed of 0m/s. When a new \textit{cmd\_vel} is computed, the controller attempts to achieve the updated speed. However, these delays can cause the AGR to stop completely, resulting in sluggish performance and difficulty in reaching the intended speed. \looseness -1

More critically, significant delays in computing speed control can lead to collision with obstacles that appear after the latest  update of speed commands based on the most recent environment perception, posing serious safety risks.
This issue becomes increasingly important as human-AGR collaboration grows, making safety a top priority. 
Conversely, if the AGR's maximum speed~$s_{max}$ is low and the CPU frequency~$f$ is high, it results in wasted computing energy without providing real benefits, leading to inefficient energy use. \looseness -1

We examine how the ($f, s_{max}$) pair influences the AGR's performance, focusing on key factors such as power consumption, end-to-end latency, and minimum reaction distance. Understanding these dynamics provides valuable insights for optimizing system behavior.  \looseness -1

\noindent{\textbf{Total Power.}} Figure \ref{fig:total_power} shows the total power consumption for various ($f, s_{max}$) pairs. We observe that the total power consumption increases with both CPU frequency $f$ and maximum speed $s_{max}$. For instance, for a fixed CPU frequency $f$, total power consumption rises by 61.67\% as maximum speed increases from 0.4m/s to 2.4m/s. Similarly, for a fixed maximum speed of 0.4m/s, the total power increases by 55.87\% when frequency rises from 1.11GHz to 2.26GHz.
Notably, at frequency 1.11GHz, the total power consumption at maximum speed 2.4m/s is lower than at 2m/s. 
This is due to increased end-to-end latency in computing the $cmd\_vel$ at lower frequencies, causing the controller to set the default speed 0m/s, as previously explained. Consequently, the average speed is 2m/s at a maximum speed of 2.4m/s and 1.8m/s at 2m/s. With this increased latency, the AGR makes more stops at maximum speed 2m/s compared to 2.4m/s, leading the AGR to consume more power to move it from a lower speed when the new speed command arrives.
\textit{These results highlight that \textbf{(Observation 1)} the power consumption and end-to-end latency are intricately linked to the CPU frequency and speed~($f, s_{max}$) pair, necessitating their joint consideration in AGR energy management.} \looseness -1 


\noindent{\textbf{End-to-End Latency (\textit{e2e} latency).}} The end-to-end (e2e) latency refers to the time taken for the computing unit to process the perceived environment and provide feedback, such as speed commands, to the motor controller. It is well known that increasing CPU frequency~$f$ lowers the e2e latency. However, Figure~\ref{fig:e2e} also illustrates that for a fixed frequency~$f$, the e2e latency increases with $s_{max}$ due to the need of navigation stack components to consider more options during planning. 
For example, at a fixed frequency~$f$ of 1.11GHz, the e2e latency increased by 222\% from~$s_{max}$ of 0.4m/s to 2.4m/s, rising from 0.36s to 1.16s, respectively. It is important to note that the AGR keeps moving at a certain average speed throughout each end-to-end latency period and can only respond to newly detected events once the period concludes. \textit{Thus, \textbf{(Observation 2)} when optimizing the AGR's frequency and max speed it is important to consider the effect of such choices on the reaction distance to provide safety guarantees during navigation}. \looseness -1

\noindent\textbf{Reaction Distance. \textit{(RD)}} The reaction distance (RD) is the distance covered while the computing unit computes the speed commands based on sensor data and is directly proportional to the e2e latency and the AGR speed. Due to the increase in e2e latency for higher  maximum speed~$s_{max}$, as shown in  Figure~\ref{fig:rd}, the reaction distance grows quadratically with the maximum speed~$s_{max}$ for a fixed frequency~$f$. For instance, for a fixed frequency of 1.11GHz, the AGR travels 2.79m at a maximum speed 2.4m/s before receiving speed command updates, which is 1828.57\% higher than the maximum reaction distance 0.14m at a maximum speed 0.4m/s (typical speed used by many studies in the evaluations, e.g.,  \cite{9303274, FAROOQ2023104285}). \textit{Given any environment, \textbf{(Observation~3)} it is desirable to ensure that the AGR never exceeds a \textit{desired maximum reaction distance} to provide safety guarantees while optimizing for energy usage.} \looseness -1

\vspace{-1.5mm}
\section{Design}
\label{sec:design}
\vspace{-2.5mm}
\subsection{System Model}
\vspace{-1mm}
Our goal is to devise an energy management algorithm that effectively utilize the given energy budget, $E_b$, for the allocated task. Given that environmental complexity affects energy consumption, it is crucial to adjust the frequency~$f$ and maximum speed~$s_{max}$ controls of the AGR based on the remaining distance to traverse~$d_r$ and the available energy~$E_{avail}$. The remaining distance~$d_r$ is estimated using the navigation path, while the available energy~$E_{avail}$ is computed from the initial budget, accounting for the energy consumed up to the current point. \looseness -1


To achieve this, we propose the Predictable Energy Consumption Controller (\textsf{PECC}) algorithm. \textsf{PECC} ensures that the AGR operates within the energy budget while maintaining optimal performance. \textsf{PECC} is flexible and can be deployed as: 1) PECC-$\delta$: This adaptive variant updates the $(f, s_{max})$ controls at each control period~$\delta$, based on current estimates of remaining distance and available energy. This approach ensures that the AGR remains responsive to changing conditions and energy estimation errors; 2) PECC: This static variant solves the optimization problem only once at the start of the task, setting the $(f, s_{max})$ controls for the entire duration. This approach is suitable for tasks with stable environments or less variability in energy consumption. \looseness -1

Next, we outline  our method for estimating the AGR's energy consumption and reaction distance for an $(f, s_{max})$ pair, which is integrated into the optimization framework.  \looseness -1

\noindent{\textbf{Power and Reactiveness Estimation Models.}} We use two independent Deep Neural Network (DNN) models to accurately estimate the power consumption and the reaction distance, respectively, of the AGR for a given $(f, s_{max})$ pair. 
We developed both models and trained them using the ADAM optimizer and mean squared error metric on the data collected from the experiments described in Section~\ref{sec:backgraound_and_motivation}.
Using separate models allows users to customize them to their specifications and independently update parameters such as power consumption or reaction distance. Additionally, we integrated these DNN models as plugins into our framework, providing users with the freedom to update them as needed.  \looseness -1

To accelerate the optimization process and leverage the plug-in nature of our framework, we utilize the DNN models to generate a comprehensive lookup table. This table estimates power consumption and reaction distance for each combination in $P = F \times S$, where $F$ represents the set of available frequencies, and $S$ denotes the set of discrete speed values, which vary with a constant difference  $\Delta_s$, ranging from a minimum speed  $S_{min}$ to a maximum speed  $S_{max}$ achievable by the AGR (0.4 m/s to 2.4 m/s for our AGR, respectively). We integrate the entries from the lookup table into the constraints of the optimization problem. 
\looseness -1

\vspace{-4mm}

\subsection{Optimization Problem} 
\label{sec:minlp}
\vspace{-0.1cm}
The optimization problem aims to maximize the utilization of the given energy budget for allocated task to enhance performance while maintaining the desired reaction distance $RD^{des}$.
To achieve this, we formulate the problem to identify the optimal pairs of frequency~$f$ and maximum speed~$s_{max}$. To expedite the optimization process, we determine the~$(f, s_{max})$ pair based on the per-meter energy budget~$E_{bpm}$. 
We calculate~$E_{bpm}$ as the ratio of the available energy budget~$E_{avail}$ (where $E_{avail}$ is initially equal to the energy budget $E_b$) to the remaining distance to traverse~$d_r$. This represents the maximum energy the AGR can use per meter.


\subsubsection{Objective}
To maximize the performance of the AGR while fully utilizing the given energy budget for the path, we define a multi-objective optimization problem:
\begin{equation}
\small
\vspace{-1mm}
    \min_{f, s_{max}, \epsilon} \left(\frac{f_{max} - f}{f_{max}} + \frac{S_{max} - s_{max}}{S_{max}}   + \omega \epsilon \right)
    \label{eq:obj}
\end{equation}
where $f_{max}$ is the maximum CPU frequency, $f$ is the selected frequency, $S_{max}$ is the AGR's maximum physical speed, $s_{max}$ is the selected speed limit while performing tasks, $\omega$ is the penalty for exceeding the per-meter energy budget $E_{bpm}$, and $\epsilon$ is  another optimization variable  representing the additional reserve energy beyond the estimated per-meter energy budget. The first two terms maximize the frequency~$f$ and speed~$s_{max}$ to improve performance in both computation and locomotion, enabling the AGR to cover more distance in less time. However, maximizing $f$ and $s_{max}$ increase energy consumption. To mitigate this, the third term introduces a penalty that balances energy usage with the given energy budget. The penalty~$\omega$, a user-defined weight, ensures that $\epsilon$ is used only when necessary, maintaining a trade-off between performance and budget compliance.
\looseness -1


\subsubsection{Constraints} PECC ensures the problem's feasibility and the AGR's safety through the following constraints.

\noindent{\textbf{Frequency and Speed Pair Selection.}}
The optimization problem finds the optimal solution among the finite set of ($f, s_{max}$) pairs $ \in P$, that minimizes the objective cost: 
\begin{equation}
\small
    (f, s_{max}) = x_{i} \cdot p_i \hspace{5mm}\forall i \in [1, |P|]
    \label{eq:sol_pair}
    \vspace{-1mm}
\end{equation}
where $p_i$ is the pair of frequency $f \in F$ and $s_{max} \in S$. $x_{i}$ is a binary auxiliary variable in set $X$, which determines the selected pair $i$. In addition, the optimization problem should select only one $(f, s_{max})$ pair: 
\begin{equation}
\vspace{-1mm}
\small
     \sum_{i=0}^{|P|} x_{i} = 1 
    \label{eq:x_constraint}
    \vspace{-1mm}
\end{equation}
Finally, the elements of the decision set $X$ are restricted to binary values.
\vspace{-1mm}
\begin{equation}
\small
\vspace{-1mm}
    x_{i} \in \{0,1\}  \hspace{5mm} \forall i \in [1, |P|]
    \label{eq:x_binary}
    \vspace{-1mm}
\end{equation}

\noindent{\textbf{Estimated Reaction Distance.}} To ensure safety, the estimated reaction  distance $RD^e$ of the $(f,s_{max})$ pair must be lower than the desired reaction distance $RD^{des}$: 
\vspace{-1mm}
\begin{equation}
\vspace{-1mm}
\small
    RD^e_{(f,s_{max})} \leq RD^{des} 
    \label{eq:reaction_distance}
    \vspace{-1mm}
\end{equation}

where, $RD^e_{(f,s_{max})}$ is the estimated RD in the lookup table for the $(f,s_{max})$ pair $\in P$. 
This ensures safety as the AGR will be able to maneuver  around the obstacles when they are perceived beyond the desired reaction distance $RD^{des}$. 

\noindent{\textbf{Estimated Energy Consumption.}} The estimated per-meter energy must not exceed the per-meter energy budget $E_{bpm}$:
\begin{equation}
\small
   P^e_{(f,s_{max})} \cdot \left(\frac{1}{s_{max}}\right) \leq  E_{bpm} + \epsilon
    \label{eq:energyBudget_const}
\end{equation}
where $P^e_{(f, s_{max})}$ is the estimated power consumption from the lookup table for the $(f, s_{max})$ pair $\in P$, and $\epsilon$ is the additional reserve energy used to ensure feasibility when $E_{bpm}$ is insufficient to obtain a feasible solution for the given $RD^{des}$. We calculate the estimated per meter energy as  the product of the time taken to traverse one meter (i.e., $1/s_{max}$) and the power required for the $(f, s_{max})$ pair. Finally, the additional reserve energy~$\epsilon$ must be greater than or equal to~0: \looseness -1
\begin{equation}
\vspace{-1mm}
\small
 \epsilon \geq 0
    \label{eq:epsilon}
    \vspace{-1mm}
\end{equation}

 \vspace{2mm}


\vspace{-1.5mm}

\subsection{Implementation}
\vspace{-1mm}

To solve the optimization problem, we use the Gurobi solver. This solver combines branch-and-bound, cutting planes, and heuristic methods to efficiently explore the solution space and obtain a near optimal solution. The optimization problem takes two inputs: ($RD^{des}, E_{bpm}$), and outputs the choosen $(f, s_{max})$ pair.
For our experiments, the solver was executed on a remote server, with an average processing time of 0.67 seconds to send the request, compute the solution, and receive the response. The data size for communication was 56 bytes. This server is a mini PC with an i7 processor, 16GB of memory, and Ubuntu~18.04 OS, connected to a WiFi network.
\textsf{PECC} could also run on the AGR computing node, depending on the compatibility of the on-board computing system and the available resources.
\looseness -1


\vspace{-1mm}
\section{Experimental Results}
\label{sec:exp_results}

\vspace{-1mm}

\noindent{\textbf{Experiment Setup.}} We conducted real-world experiments with our AGR (Figure~\ref{fig:robot}) on the map shown in Figure~\ref{fig:real_map}. For these experiments, we ran trials on the paths AB, BD, and DA with 0.5m and 0.75m RD and low, medium, and high energy budgets. Along with varying RD and energy budgets, we introduced 9 additional obstacles (see example in Figure~\ref{fig:agr_on_path}) not present in the initial map shown in Figure~\ref{fig:real_map}. The obstacle configuration were consistent across different setups for the baseline and PECC, allowing for a direct comparison between them. For the real experiments, we utilized the PECC-0 variant, where the optimal ($f, s_{max}$) pair is set at the beginning of the task. \looseness=-1

For simulations, we employed Gazebo with a 1:1 scale model of our AGR on the environment map shown in Figure~\ref{fig:sim_map}. We obtained these results for the PT, QU, and RS paths,  using 0.25m, 0.5m, and 0.75m RD, for  low, medium, and high energy budgets. Including PECC-0, we tested another variant PECC-$\delta$ of our approach with the baseline, where $\delta = 10$, indicating that the ($f, s_{max}$) pair was updated every 10 seconds based on available energy for the remainder of the path. To ensure consistency between the real-world experiments and simulations, we introduced the e2e latency by simulating similar experiments for different ($f, s_{max}$) pairs  $\in P$ and creating a lookup table. We then calculate the non-negative difference between the AGR's and simulations e2e latency for the elected ($f,s_{max}$) pair and applied it as delay between the computed speed command and the locomotion motor controller.
\looseness=-1 


\noindent{\textbf{Baseline.}} We utilize the \emph{Energy-Efficient} (EE) baseline for comparison with our approach. EE focuses on minimizing per meter energy consumption as presented in \cite{eemrc}. We implemented EE in Gurobi to find the near  optimal energy-efficient ($f, s_{max}$) pair while ensuring the estimated reaction distance, $RD^e$, satisfies Equation~\ref{eq:reaction_distance}. \looseness=-1


\noindent{\textbf{Performance Metrics.}} We used the following performance metrics: 1) \emph{Relative Travel Time} (RTT): The ratio of the time taken by the AGR to reach the goal using EE compared to PECC (i.e., travel time of EE/PECC). A RTT greater than 1 indicates lower travel time than EE, while a value below 1 reflects longer travel time than EE; 2) \emph{Energy Budget Usage} (EBU): The ratio of total energy consumed by the AGR to travel a path to the given energy budget, ideally close to 1.

\vspace{-1mm}
\subsection{Performance Results}
\begin{figure}[!t]
  \centering

        \subfloat[]{
  
  \includegraphics[width=0.4
  \textwidth]{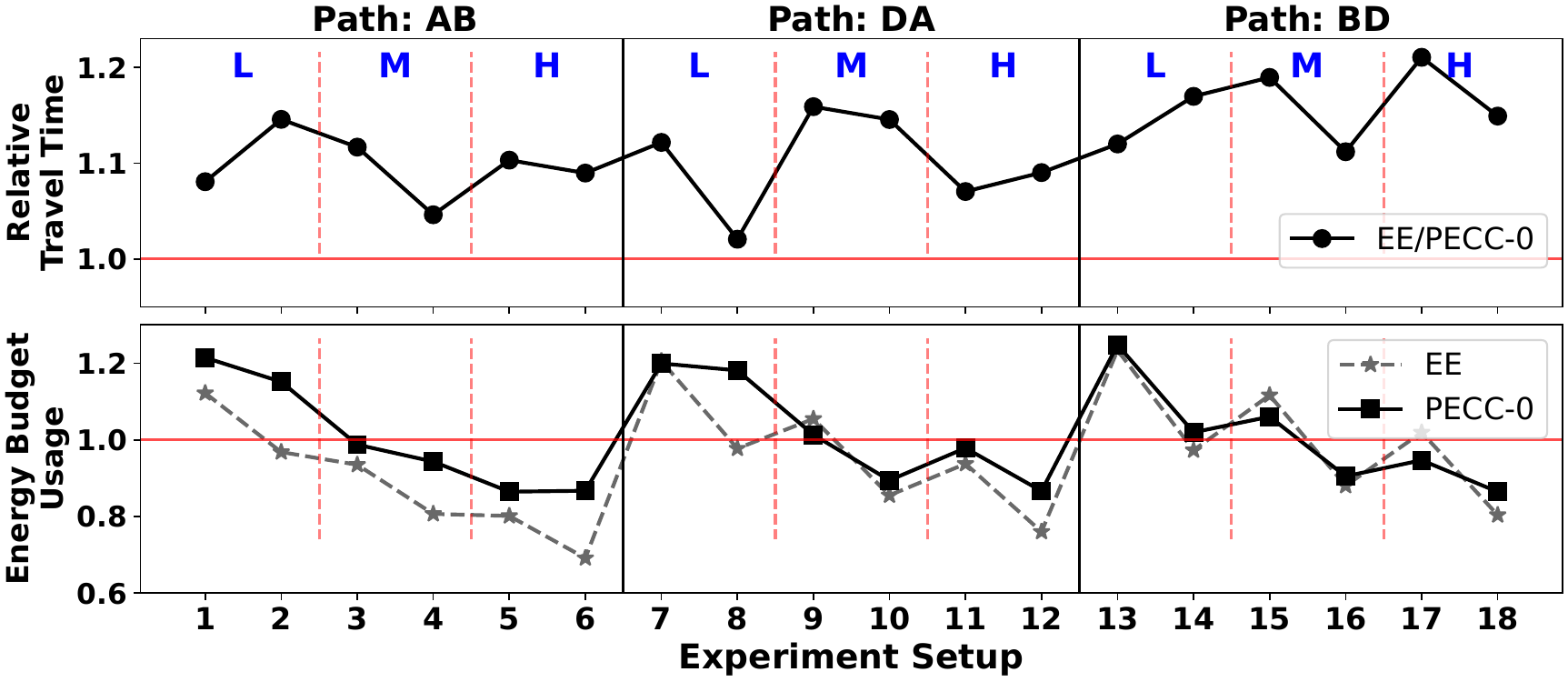}
  \label{fig:act_com_results}
  }
\vspace{-2mm}
        \subfloat[]{
  \includegraphics[width=0.4\textwidth]{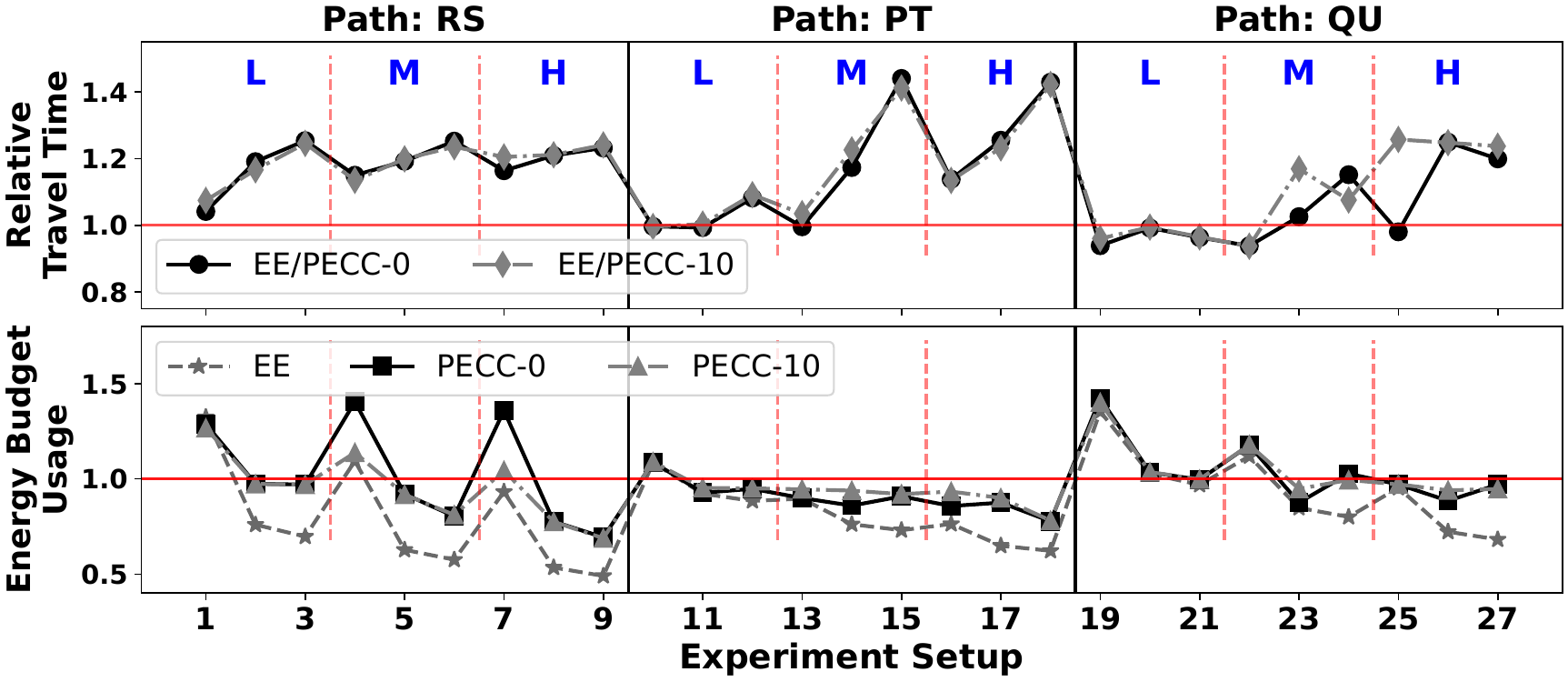} 
  
  \label{fig:sim_com_results}
  }
\vspace{-2mm}
  \caption{Comparison of relative travel time and energy budget usage between \textsf{EE} and \textsf{PECC} across different reaction distances,  energy budgets and paths for (a) real-world experiments using the map shown in  Figure~\ref{fig:real_map}, and (b) simulations using  the map shown in  Figure~\ref{fig:sim_map}.  \looseness -1
  }
  \label{fig:combined_results}
  \vspace{-0.7cm}
\end{figure}

Figure~\ref{fig:combined_results} shows the performance of both real-world and simulated results with the performance metrics. In Figures~\ref{fig:act_com_results} and~\ref{fig:sim_com_results}, the top and bottom plots shows RTT, and EBU, respectively. Each plot is divided into three columns, separated by black vertical lines representing different paths. In each column the setups are arranged from left to right by increasing energy budgets, separated by red vertical lines: low (L), medium (M), and high (H). Within each energy budget we show RD in increasing order. Specifically, Figure~\ref{fig:act_com_results} represents RD 0.5m and 0.75m, while Figure~\ref{fig:sim_com_results} covers RD 0.25m, 0.5m, and 0.75m. 
Table \ref{table1} shows a comparison of travel time between real-world and simulation experiments for 0.5m and 0.75m RD, and for low, medium, and high energy budgets. For performance comparison, we compare the average of the best 3 runs for each setup. \looseness -1

\noindent{\textbf{Real-World Results.}} We conducted 10 trials for each of the 2 RDs, 3 energy budgets, and 3 paths, totaling 180 runs. On average, each run took 122s, amounting to approximately 6 hours of AGR travel time. These experiments with AGRs demonstrate that utilizing the given energy budget enhance performance in terms of travel time. For instance, across all setups, PECC outperforms EE with an average RRT of 1.12, meaning on average PECC reaches the destination 11.2\% faster than EE while exceeding the energy budget by only 1\%. The best result is achieved in setup 17 on path BD for high energy budget and 0.5m RD, where PECC is 17.39\% faster than EE and consumes 95\% of the given energy budget. 
We observe that for insufficient energy budgets, both EE and PECC exceed the allocated energy (setups: 1, 7, and 13). In these cases, while PECC consumes slightly more energy, it achieves a higher RTT. For instance, on path AB with RD 0.5m and a low energy budget, PECC is 7.5\% faster than EE while consuming 8.25\% higher energy. This increase in energy consumption can be controlled by adjusting the weight, $\omega$, in the objective function~(Equation~\ref{eq:obj}). Given a sufficient energy budget, PECC stays within the limit for lower travel time. For instance, on path DA with an RD 0.5m and a high energy budget, PECC consumes 98\% of the budget and is 6.6\% faster than EE. This is because a higher energy budget allows PECC to select a higher CPU frequency $f$ and maximum speed $s_{max}$. \looseness=-1

Similar to the effect of increasing the energy budget, a larger RD allows higher speeds. This is because higher RD unlocks higher speed for the same frequency. 
Increased speed reduces travel time  (shown in Table \ref{table1}),  which in turn lowers the overall energy consumption. Therefore, for low budgets specifically, we observe that energy consumption decreases with increased RD. 
\looseness=-1

\noindent{\textbf{Simulation Results.}} For simulations, we conducted 25 trials for 3 different RDs, 3 energy budgets, and 3 paths, totaling 675 runs. On average, each run took 125 seconds, amounting to approximately 24 hours of simulation time. Table \ref{table1} shows the comparison of travel time trends based on RD and energy budget. The simulations exhibit similar trends to those observed in the real-world experiments, with increasing RD leading to reduced travel time.  \looseness-1

Figure~\ref{fig:sim_com_results} shows that the simulation results, similar to the real-world experiments, exhibit a higher RTT of 1.13 for PECC-0 and 1.15 for PECC-10, making them 10.6\% and 12.15\% faster than EE, respectively. The average EBU for EE, PECC-0, and PECC-10 are 0.85, 0.99, and 0.98, respectively. EE being energy efficient consumes only 85\% of the energy budget leading to 
 13.44\% and 15.23\% longer travel time than PECC-0 and PECC-10, respectively. 
In most setups, RTT increases with RD and energy budget, as both factors allow higher speeds while meeting the RD constraint in  Equation~\ref{eq:reaction_distance}. 
For example, on path PT with medium energy budget, the RTT increases from 1 to 1.44 (which is 30.58\% faster than EE) as RD grows from 0.25m to 0.75m, representing a 44\% increase in RTT. Similarly, for the same path and RD, RTT rises by 14\% when the energy budget increases from low to high.  
\looseness=-1

We also observe that, PECC-10 consistently achieves the best RTT while maintaining a similar EBU as PECC-0 across setups. This is because PECC-10 dynamically adjusts the ($f, s_{max}$) pair based on available energy, which lowers travel time with similar EBU to PECC-0. In some cases (e.g., on path RS), it even achieves similar travel time while remaining much closer to the given energy budget.
\looseness=-1

\begin{table}[!t]
\caption{Comparison of travel time (in seconds) between real-world and simulated results, for increasing reaction distance (RD, in meters) and energy budgets ($E_b$)}
\vspace{-2mm}
    \centering
    \resizebox{\columnwidth}{!}{%
     \begin{tabular}{|>{\centering\arraybackslash}p{0.25cm}|>{\centering\arraybackslash}p{0.6cm}|>{\centering\arraybackslash}p{2.3cm}|>{\centering\arraybackslash}p{2.3cm}|>{\centering\arraybackslash}p{0.6cm}|>{\centering\arraybackslash}p{2.3cm}|>{\centering\arraybackslash}p{2.3cm}|}
    \hline
        \multirow{2}{6em}{\textbf{$E_b$}}  & \multicolumn{3}{|c|}{\textbf{Real World Experiments (EE/PECC-0)}} & \multicolumn{3}{c|}{\textbf{Simulation Experiments (EE/PECC-0)}} \\ \cline{2-7} 
        & \textbf{Path} & \textbf{RD = 0.5} & \textbf{RD = 0.75} & \textbf{Path} & \textbf{RD = 0.5} & \textbf{RD = 0.75} \\ \hline
        \multirow{3}{6em}{\rotatebox[origin=c]{90}{\textbf{Low}}} &  AB & 81.62 / \textbf{75.53} & 79.11 / \textbf{69.04} & RS & 86.97 / \textbf{73.03 } & 74.75 / \textbf{59.61} \\ 
        &  DA & 113.97 / \textbf{101.6} & 100.16 / \textbf{98.13} & PT & 101.37 / \textbf{102.07} & 93.12 / \textbf{86.05} \\ 
        & BD & 157.67 /\textbf{ 140.71} & 130.23 / \textbf{111.33} & QU & 114.08 / \textbf{115.0} & 102.16 / \textbf{106.03}\\ \hline
       \multirow{3}{8em}{\rotatebox[origin=c]{90}{\textbf{Med}}}  &  AB & 81.62 / \textbf{73.09} & 79.11 / \textbf{75.72} & RS & 86.97 / \textbf{72.94} & 74.75 / \textbf{59.68}\\ 
       &  DA & 113.97 / \textbf{98.33} & 100.16 / \textbf{87.43} & PT & 101.37 / \textbf{86.38} & 93.12 / \textbf{64.64}\\ 
       &  BD & 157.67 / \textbf{132.49} & 130.23 / \textbf{117.11} & QU & 114.08 /\textbf{ 111.2} & 102.16 / \textbf{88.69}\\ \hline
      \multirow{3}{6em}{\rotatebox[origin=c]{90}{\textbf{High}}} &  AB & 81.62 / \textbf{73.99} & 79.11 / \textbf{72.6} & RS & 86.97 / \textbf{71.91} & 74.75 / \textbf{60.67} \\ 
       &  DA & 113.97 / \textbf{106.47} & 100.16 / \textbf{91.87 }& PT & 101.37 / \textbf{80.76} & 93.12 / \textbf{65.15} \\ 
       &  BD & 157.67 / \textbf{130.2} & 130.23 / \textbf{113.32} & QU & 114.08 / \textbf{91.31} & 102.16 / \textbf{85.18} \\ \hline
    \end{tabular}
    }
    \label{table1}
    
    \vspace{-0.6cm}
\end{table}

\vspace{-1mm}
\subsection{Discussion and Future Work}
\vspace{-1mm}
Our experimental results demonstrate that increasing RD allows for higher speeds and reduced travel time. However, there is a trade-off: beyond a certain RD, the risk of the AGR colliding with obstacles increases. We plan to identify the optimal RD for a given budget to effectively balance performance and safety. \looseness -1

Furthermore, simulation results for PECC-10 reveal that dynamically adjusting the ($f, s_{max}$) pair leads to improved travel time and more efficient energy budget utilization. Moving forward, we will explore methods for redistributing energy along the remaining distance to further enhance energy efficiency and overall system performance. We will also develop polynomial-time approximation algorithms to further reduce computation time. \looseness -1

\vspace{-1mm}
\section{Related Work}
\vspace{-1mm}
\label{sec:related}

Research focused on managing limited energy availability on AGR can be can be grouped into the following categories:

\noindent{\textbf{Energy Budgeting.}} Several studies propose task and charging scheduling strategies for AGRs by budgeting energy for specific tasks or periods~\cite{tcm, 9981285, 9636815, Boccia_Masone_Sterle_Murino_2023, Chen_Xie_2022, 7294146, McGowen_Dagli_Dantam_Belviranli_2024, tcm_m}. These solutions often lack control over energy utilization, necessitating replanning based on updated energy availability. Tasks such as food delivery, which cannot be easily rescheduled once started. Additionally, certain tasks may lead to inefficient energy use, such as operating at maximum computing frequency unnecessarily. This highlights the need for more effective energy management strategies.

\noindent{\textbf{Energy-Efficient Solutions.}} Studies  on energy-efficient solutions focus on minimize  locomotion and computation energy consumption~\cite{Lamini_Benhlima_Elbekri_2018, Kim_Kim_2014, Li_Wang_Chen_Kan_Yu_2023, Di_Franco_Buttazzo_2015, 6385568,e2m, Rossi_Vaquero_Sanchez_Net_da_Silva_Vander_Hook_2020, Tang_Shah_Michmizos_2019, eemrc}. However, these approaches often prioritize energy efficiency over maximizing the energy budget for improved performance.

These studies often work independently, limiting their potential by not combining both approaches. \textit{To the best of our knowledge, no study focuses on maximizing the energy budget utilization for scheduled tasks to enhance performance while guaranteeing reactiveness by design (through reaction distance).} \looseness -1

\vspace{-1mm}
\section{CONCLUSIONS}
\label{sec:conclusion}
This study identifies the limitations of existing approaches that either focus on energy efficiency or task allocation without integrating both strategies effectively. We first analyze AGR performance across various CPU frequencies and maximum speeds for power consumption and end-to-end latency. We then introduce the Predictable Energy Consumption Controller (PECC), designed to maximize energy budget utilization for improved performance. Experimental results with a real AGR shows a 17\% reduction in travel time compared to the energy-efficient baseline, using 95\% of the energy budget. Simulations further demonstrate a 31\% reduction in travel time with PECC, utilizing 91\% of the energy budget.
\looseness -1

\bibliographystyle{IEEEtran}
\bibliography{IEEEabrv,biblography}

\end{document}